\pdfoutput=1

\documentclass[11pt]{article}

\usepackage{acl}

\usepackage{times}
\usepackage{latexsym}

\usepackage[T1]{fontenc}

\usepackage[utf8]{inputenc}

\usepackage{microtype}

\usepackage{inconsolata}

\usepackage{graphicx}
\usepackage{amsmath}
\usepackage{amsfonts}
\usepackage{makecell}
\usepackage{tabularx}
\usepackage{float}
\usepackage{multirow}
\usepackage{booktabs}
\usepackage[justification=centering]{caption}
\usepackage[justification=centering]{subcaption}
\title{Syllable-level lyrics generation from melody exploiting character-level language model}

\author{%
    Zhe Zhang\textsuperscript{1}, 
    Karol Lasocki \textsuperscript{2}\textsuperscript{†}, 
    Yi Yu\textsuperscript{1}\textsuperscript{*}, 
    Atsuhiro Takasu\textsuperscript{1} \\  
    National Institute of Informatics, SOKENDAI\textsuperscript{1} \\
    Aalto University\textsuperscript{2}\\
   \texttt{\{zhe, yiyu, takasu\}@nii.ac.jp, karolasocki@gmail.com} \\
}

\begin{document}
\maketitle
\begingroup\def\thefootnote{*}\footnotetext{Yi Yu is the corresponding author.}\endgroup
\begingroup\def\thefootnote{†}\footnotetext{Karol was involved in this work during the internship at National Institute of Informatics (NII), Tokyo.}\endgroup

\begin{abstract}
The generation of lyrics tightly connected to accompanying melodies involves establishing a mapping between musical notes and syllables of lyrics. This process requires a deep understanding of music constraints and semantic patterns at syllable-level, word-level, and sentence-level semantic meanings. However, pre-trained language models specifically designed at the syllable level are publicly unavailable. To solve these challenging issues, we propose to exploit fine-tuning character-level language models for syllable-level lyrics generation from symbolic melody. In particular, our method endeavors to incorporate linguistic knowledge of the language model into the beam search process of a syllable-level Transformer generator network. Additionally, by exploring ChatGPT-based evaluation for generated lyrics, along with human subjective evaluation, we demonstrate that our approach enhances the coherence and correctness of the generated lyrics, eliminating the need to train expensive new language models.
\end{abstract}

\section{Introduction}

Generating lyrics from a given melody is a subjective and creativity-driven process that does not have a definitive correct answer. Recognizing the importance of subjective and creativity-driven generation processes is essential for advancing the development of AI. By embracing and enabling such processes, we can pave the way for more nuanced and expressive AI-generated lyrics. Accordingly, evaluating the quality of subjectively and creativity-driven generated lyrics has become a fascinating topic. Our system focuses on generating lyrics from symbolic melodies and could serve as a valuable creative aid, collaborating with artists throughout the entire songwriting process. The use of symbolic melodies allows for effortless and frequent modifications, facilitating iterative creative exploration.

In this work, we explore the generation of lyrics from simplified symbolic melodies consisting of 20 notes. Our aim is to maintain the alignment between the syllables of the lyrics and the corresponding melody notes during the inference stage. To achieve this, we propose a melody-encoder-syllable-decoder Transformer architecture, which generates syllables sequentially in accordance with the melody. However, due to the scarcity of paired lyrics-melody data available for training, this approach could lead to producing lyrics that are not coherent and grammatically not correct, such as ``\textit{you gotta o in what the you used to life}''.

The dataset we are using is described in \cite{yu_conditional_2021}, and it only contains approximately 10,000 paired lyrics-melody sequences. Each lyrics sequence in the dataset contains 20 syllables in length, and there may be samples where syllables are occasionally missing due to misalignment, or lack of corresponding notes. These problems significantly hinder the training of a model to comprehend and generate coherent language.

On the other hand, due to the constraint of syllable-level generation, it is difficult to directly apply pre-trained language models that already have an understanding of linguistic knowledge, due to the scarcity of syllable-level language models. The utilization of the widely popular word-piece encoding is not feasible in our task because one word consists of different numbers of syllables. This would potentially affect the probabilities of generating multi-syllable words. A possible alternative approach to train a custom language model at the syllable level is using a large, clean text corpus that has been segmented into syllable-level texts, which can then be fine-tuned specifically for the task of generating lyrics, but it is also difficult to construct such kind of dataset. Another solution is to fine-tune a character-level language model, refining it to generate syllable sequences. In this work, we focus on the latter approach, which aims to fine-tune a character-level language model for re-ranking the candidates generated by a melody-encoder-syllable-decoder Transformer \cite{vaswani2017attention}.

We take inspiration from the usage of language models in re-ranking speech recognition token candidates \cite{Bhler2005UsingLM}. Considering the sentence ``\textit{Last \textbf{x} was windy}'', and the speech recognition system candidates \textit{knight} and \textit{night}. Due to the pronunciation similarities, the word \textit{knight} could be given a higher probability when recognizing speech. However, a language model would easily fix the mistake, assigning a higher probability to the word \textit{night} instead.

Another inspiring work by \citet{wang2021modeling} focused on video comment generation tasks, In this work, the probability of previous text token, the probability of future text token, and the mutual dependency between comment texts and video are modeled by three separately trained neural networks. The probabilities from all three models are then combined and the best candidate from the main comment generation Transformer model is selected, improving coherence and relation between comments and video. 

In our study, using a real example from our models, given the sentence ``\textit{you gotta}'', the lyrics generation model could predict possible next tokens as \textit{o} rather than \textit{treat} because of the limited training data it learned from, but it is neither grammatically correct nor semantically meaningful. In this case, a powerful language model would know that the latter is more likely to form a coherent sentence. Using a fine-tuned language model to refine the semantic meanings within generated syllable-level lyrics, we are able to improve the generated sequence from ``\textit{you gotta o in what the you used to life}'' to ``\textit{you gotta treat me to maybe understand you}''. As one phrase of lyrics, the revised sequence is much more coherent and interesting than the original version.

The main contributions of this work can be summarized as follows:
\begin{enumerate}
    \item Training a melody-encoder-syllable-decoder Transformer model to generate lyrics syllable by syllable, ensuring semantic correlation with individual notes in the melody.
    \item Proposing exploiting the fine-tuned character-level pre-trained language models for refining candidate syllables generated by the Transformer decoder to ensure the coherence and correctness in the generated lyrics, overcoming the difficulty of unavailable pre-trained syllable-level language models.
    \item Designing a beam search and re-ranking technique to integrate the fine-tuned language model with the Transformer decoder to predict re-ranked lyrics candidates.
\end{enumerate}

\section{Proposed methods}

By exploiting fine-tuning a pre-trained language model, we have successfully designed syllable-level lyrics generation architecture from symbolic melody exploiting character-level language model depicted in \autoref{fig:mutual}. In this section, we will introduce the details of the proposed methods.

\subsection{Syllable-level lyrics generation from melody}

\begin{figure*}[htb]
    \centering
    \includegraphics[width=0.75\textwidth]{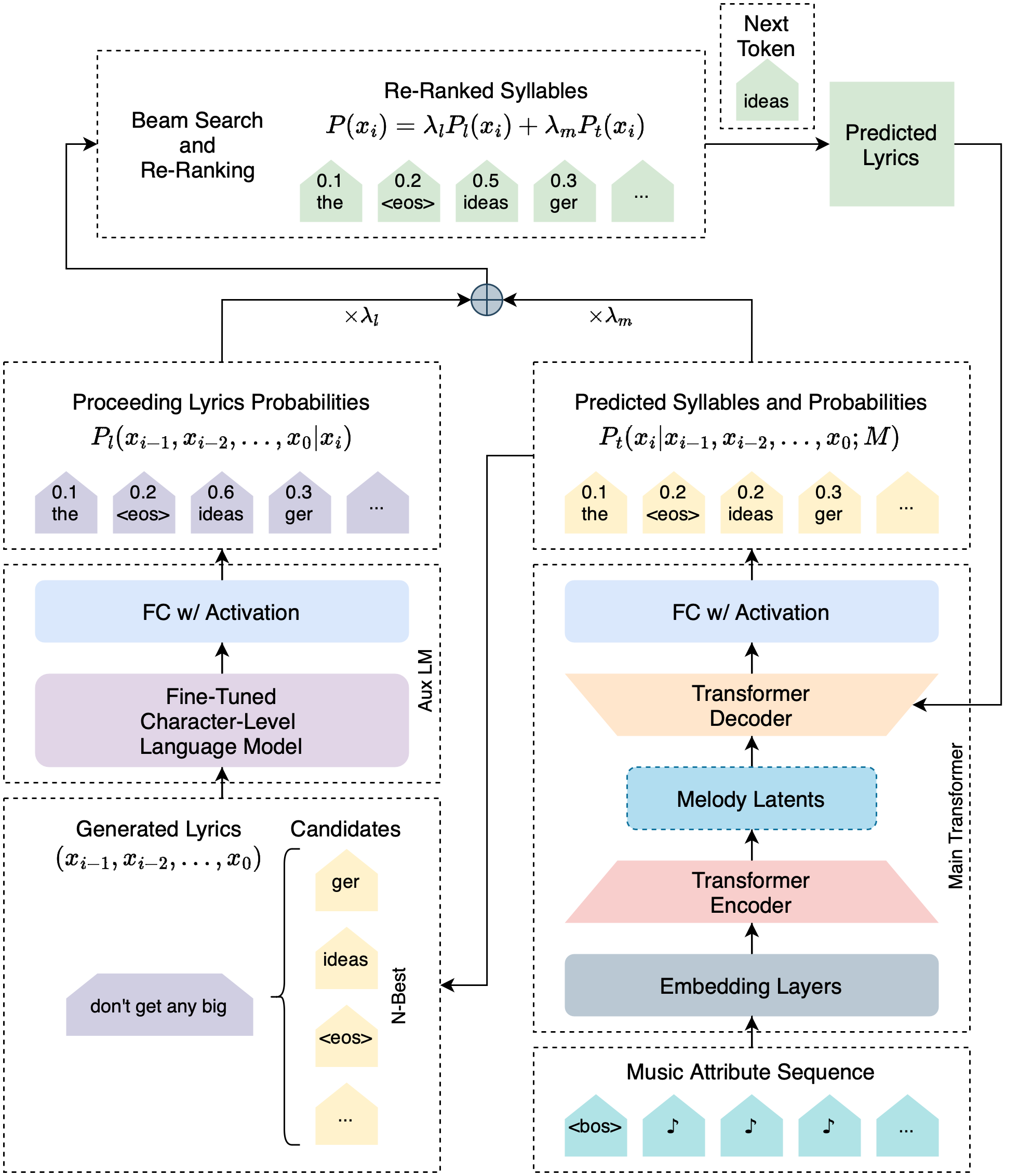}
    \caption{Transformer-based melody-encoder-syllable-decoder architecture exploiting character-level language model. }
    \label{fig:mutual}
\end{figure*}

As shown in \autoref{fig:mutual}, the Transformer on the right side generates the candidate syllable tokens based on the encoded melody latent representations $M$ and previously generated lyrics. The fine-tuned language model on the left evaluates the probability of the candidates based on the given lyrics generated, which aims to improve the coherence and correctness of the generated lyrics. 

As an example shown in \autoref{fig:mutual}, the proposed model has generated a sequence of lyrics tokens \textit{don't get any big} in previous time steps. In the current time step, the Transformer decoder predicts syllable \textit{ger} with a probability of 0.3 and predicts syllable \textit{ideas} with a probability of 0.2. Considering the Transformer is trained on a limited amount of data, it might assign a higher probability to \textit{ger} because the syllables can construct a word \textit{bigger}. However, the language model, which is trained on a large amount of corpus, can predict \textit{ideas} with a higher probability of 0.6 because the sentence \textit{don't get any big ideas} is more meaningful in natural language. Then, in the re-ranking stage, token \textit{ideas} can be assigned the highest probability after weighting the two probabilities. In such a way, the language model can help the Transformer generator predict better lyrics in terms of grammar and meaning.

We focus on exploiting the language model in \autoref{fig:mutual}, hoping to improve the coherence and correctness of the lyrics generated by the main model by using the knowledge of a pre-trained character-level language model to re-evaluate the token probabilities during beam search generation. It could improve the results and generated lyrics quality as opposed to using solely the baseline encoder-decoder Transformer model. 

The probability that the language model computes would be $P_l(x_{i-1}, x_{i-2}, …, x_0| x_i)$, where $x_i$ is the \textit{ith} syllable of the lyrics. We only start using the language model from the second generation step, ensuring that $x_0$ is known.
The probability modelled by the Transformer model would be $P_m( x_i|x_{i-1}, x_{i-2}, …, x_0, f_n, f_{n-1}, …, f_0)$, where $f_i$ are the melody features at time $i$. 

The total probability for a given token is then:
$$P(x_i) = \lambda _l * P_l(x_i) + \lambda _t * P_t(x_i), $$ where $\lambda _l + \lambda _t = 1$ are weights indicating which model we prioritize. 

In our work, we fine-tune the pre-trained Google CANINE \cite{Clark_2022} model using our dataset. We chose CANINE as it is a widely recognized open-source character-level language model. We use the task of Next Sentence Prediction (NSP), i.e., given a syllable $s$ and a lyric $l$, predicting the probability $P(s | l)$ that $s$ follows $l$. Note that in the case of character-level language models, both $s$ and $l$ are sequences, hence the NSP approach can work well. Fine-tuning is essential since the word distribution of lyrics differs significantly from that of resources typically used in training the language model, such as books or Wikipedia. For instance, lyrics contain the words \textit{love}, \textit{hate}, and \textit{gotta} more frequently, and have more lenient grammar.

\subsection{Dataset for fine-tuning the language model}

We have created the dataset for fine-tuning CANINE based on our lyrics dataset \cite{yu_conditional_2021}. As each syllable of lyrics with its preceding sequence in our original dataset can be thought of as a data point, we are able to obtain a fine-tuning dataset of a considerable size of over 2 million examples. 

An example of constructing data samples can be seen in \autoref{tab:dataset}. For negative examples (label 0), we select a random syllable from the same lyrics sequence that is not the correct continuation of the input sequence. We believe that using syllables from the same lyrics sequence poses a bigger challenge to the language model compared with selecting from the whole vocabulary since the syllables in the same sequence are more plausible candidates than unrelated ones from the vocabulary. 

\begin{table*}[htb]
\small
\centering
\begin{tabular}{cc|c}
\toprule
lyrics input & candidate syllable & label \\
\midrule
i know why your mean to me when & \textcolor{green}{\_}i & 1 \\
i know why your mean to me when & \textcolor{green}{\_}the & 0 \\
i \textcolor{red}{e} why your mean to me when & \textcolor{green}{\_}i & 0 \\
i know why your mean to me when & i & 0 \\
i know why your mean to me when i & \textcolor{green}{\_}call & 1 \\
i know why your mean to me when i & \textcolor{green}{\_}on & 0 \\
i know why\textcolor{red}{tel} mean to me when i & \textcolor{green}{\_}call & 0 \\
i know why your mean to me when i & call & 0 \\
… & & \\
i know why your mean to me when i call on the & \textcolor{green}{\_}tel & 1 \\
i know why your mean to me when i call on the & \textcolor{green}{\_}the & 0 \\
i know why your mean to\textcolor{red}{<eos>}when i call on the & \textcolor{green}{\_}tel & 0 \\
i know why your mean to me when i call on the & tel & 0 \\
i know why your mean to me when i call on the tel & e & 1 \\
i know why your mean to me when i call on the tel & \textcolor{green}{\_}when & 0 \\
i know why your mean to me when i call on\textcolor{red}{e} tel & e & 0 \\
i know why your mean to me when i call on the tel & \textcolor{red}{\_}e & 0 \\
i know why your mean to me when i call on the tele & phone & 1 \\
i know why your mean to me when i call on the tele & <eos> & 0 \\
i know why your mean to me when i call on the tele & \textcolor{red}{\_}phone & 0 \\
i know why your mean to me when i call on the telephone & <eos> & 1 \\
i know why your mean to me when i call on the telephone & \textcolor{green}{\_}phone & 0 \\ 
\bottomrule
\end{tabular}
    \caption{An example of how the fine-tuning dataset is built from sequences of lyrics. The reasons for negative labels are marked in red, while correct spaces are highlighted in green.}
    \label{tab:dataset}
\end{table*}

Since the syllables are separated by blank spaces in the melody-lyrics dataset, the lyrics it generates are different from the correctly formatted language that CANINE is used to. Therefore, to enable the pre-trained CANINE model to learn the blank space distribution, we introduce negative data samples with incorrect spacing, i.e., some without the space like ``\textit{example}'' and some with it like ``\textit{\_example}''. The more probable variant is selected and used to form the context for the next generation step. This allows us to use the language model for connecting the syllables generated by the Transformer into full words. Specifically, for the first three predictions of the NSP task, we introduce negative examples with incorrect spacing, and in the following predictions, we set an incorrect spacing probability of 60\%, to avoid significantly increasing the size of the dataset. Negative examples with random syllables selected as the candidate have the spacing information preserved from the original location of the candidate. For instance, in the example \textit{``i know why your mean to me when i call on the'',  ``\_the''}, the candidate syllable \textit{the} has a space in front of it, since this is how it originally appeared in the lyric.

Moreover, in order to improve the robustness of the model and its ability to recover from mistakes, in 40\% cases we also include examples where one syllable from the preceding lyrics is randomly switched to a different syllable from the same lyric. For instance, in \textit{``i know whytel mean to me when i'', ``\_call''}, the syllable \textit{tel} has randomly replaced the syllable \textit{\_your}, making it a negative data sample. Since we are aiming to simulate mistakes, we randomly insert a space before the syllable with a probability 50\%.

The dataset used for training the model is imbalanced, with a higher proportion of negative examples compared to positive examples. The reason for such construction is that it reflects the real-world scenario, where the model performs a beam search with multiple candidates, out of which only one is expected to be correct. The model is able to perform well despite the imbalances, achieving convergence after 5 epochs of training.


\subsection{Beam search and re-ranking}
\label{sec:beam}

At each beam search step excluding the first, we have $n = \mathrm{beam \ size}$ candidate syllables for each of the $n$ beam sequences with the highest probabilities: $S = s_1, ..., s_n$, in total $n \times n$ candidate sequences to consider. The generated candidate syllables are then 
$$
G = g_{1, 1}, g_{1, 2}, ..., g_{1, n}, g_{2, 1}, ... , g_{n, n}.
$$

At the first beam search step, we start with a single \textit{<BOS>} (beginning of sentence) special token, and generate the $n$ best candidates for it, which become $s^0 = s_1, ..., s_n$.

Each generated candidate is associated with the probability assigned by the main transformer model  $\mathbf{M} \in \mathbb{R}^{n \times n}$. We also compute the fine-tuned language model probabilities for the sequences 
$$
L_{i, j} = lm(s_i, g_{i, j}).
$$

The final combined probabilities are then
$$
C_{i, j}^t = \lambda_m * M_{i, j} + \lambda_l * L_{i, j},
$$
for $0 < i,j \leq n$ at each timestep $t \in T$, where $\lambda_m + \lambda_l = 1$ are weights assigned to the predictions of each model. We then select the $n$ best sequences, and continue the process using them as the new $s^{t+1} = s_1, ..., s_n$.

However, this does not take into account the probabilities at previous timesteps. If we consider text generation, the sequence ``\textit{I am co ming home}'' might receive a low score, since \textit{home} is just one of the possible continuations where one can be \textit{coming}. However, the sequence ``\textit{the the could ath lete}'', despite making less sense, could score higher, this is because having predicted syllable ``\textit{ath}'', the model would be highly confident that the next syllable is ``\textit{lete}''. 

To prevent that, a standard technique is to compare the candidates using cumulative probabilities, given by
$$
C^t_{i, j} =  \lambda_m * M^t_{i, j} + \lambda_l * L^t_{i, j} + C^{t-1}_{i, j},
$$
where $C^0_{i, j} = M^0_{i, j}$, since we do not engage the language model in the first beam search step.

\section{Experiments}

\subsection{Experiment setup}

We trained a melody-to-lyrics Transformer model as a strong baseline and the basis of our methods. To leverage the ability of the language model, we set the weight of the fine-tuned language model to 75\%, leaving 25\% for the Transformer. Although the use of the language model noticeably slows down the beam search procedure, a complete evaluation on a validation set containing approximately 1000 examples can still be done in less than 3 hours on an A100 GPU. The fine-tuning of the language model was performed using default hyperparameters from the huggingface library \cite{huggingface-https://doi.org/10.48550/arxiv.1910.03771}, and lasts less than one day on an A100 GPU, despite the size of the fine-tuning dataset.

\subsection{Objective metrics}
\label{sec:obj}

Evaluating creative text objectively is an exceedingly challenging task. Sequence evaluation metrics such as ROUGE and BLEU have limited utility when evaluating creative text because they mainly focus on measuring n-gram similarities between generated sequences and reference sequences. When evaluating creative text, it is crucial to understand that the goal is not to replicate a single ground truth reference. In some cases, an outstanding lyric may be unfairly penalized simply because it deviates from the ground truth, despite effectively fitting the melody and showcasing artistic excellence.

To the best of our knowledge, there exists no objective metrics that can comprehensively capture the quality of the generated lyrics. Therefore, we only use the objective metrics as a means to validate the reconstruction ability of the proposed model. 

\begin{table*}[htb]
\centering
\small
\begin{tabular}{c|c|c|c}
\toprule
Metric & SDN\cite{duan_semantic_2023} & Transformer & Transformer + LM \\ 
\midrule
\makecell{ROUGE F score \\ (1,2,L)} & 0.1301, 0.0008, 0.0981 & 0.1476, 0.0354, 0.1248 & 0.1439, 0.0289, 0.1186 \\
\midrule
\makecell{Sentence BLEU \\ (2,3,4-gram)} & 0.0171, 0.0074, 0.0049, & 0.0637, 0.0454, 0.0374 & 0.0576, 0.0386, 0.0308 \\ 
\midrule
\makecell{BERT Scores \\ (Precision, Recall, F1)} & 0.8771, 0.8870, 0.8819 & 0.967, 0.968,
0.967 & 0.967, 0.969, 0.968 \\
\bottomrule
\end{tabular}
    \caption{Objective metrics on the validation dataset}
    \label{tab:results}
\end{table*}

\begin{figure*}[htb]
     \centering
     \begin{subfigure}[b]{0.9\textwidth}
        \centering
         \includegraphics[width=0.9\textwidth]{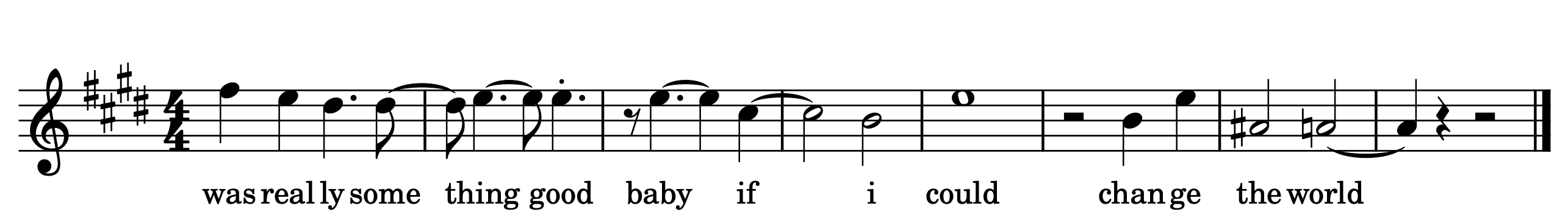}
         \caption{Ground-truth lyrics.}
         \label{fig:sheet_gt}
     \end{subfigure}
     \hfill
     \begin{subfigure}[b]{0.9\textwidth}
         \centering
         \includegraphics[width=0.9\textwidth]{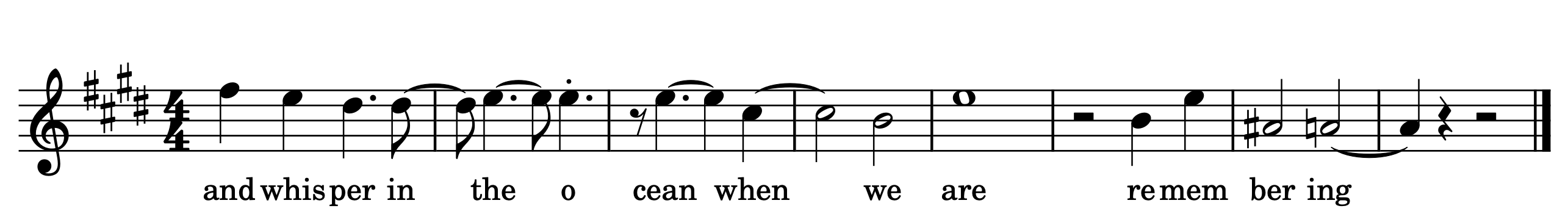}
         \caption{Generated lyrics by Transformer.}
         \label{fig:sheet_baseline}
     \end{subfigure}
     \hfill
     \begin{subfigure}[b]{0.9\textwidth}
         \centering
         \includegraphics[width=0.9\textwidth]{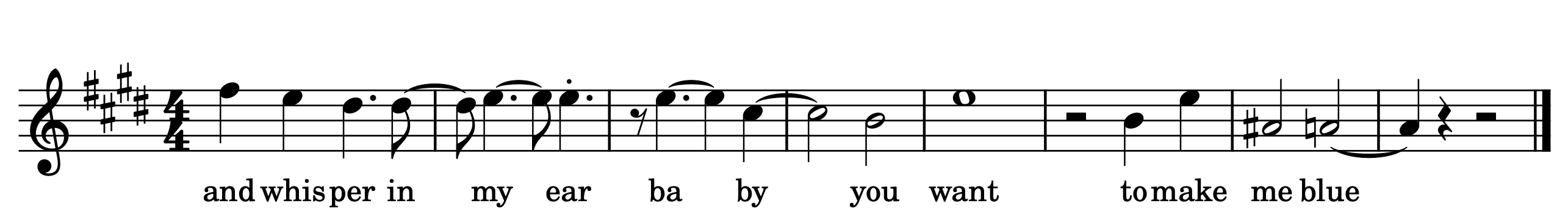}
         \caption{Generated lyrics by Transformer + LM.}
         \label{fig:sheet_ours}
     \end{subfigure}
        \caption{Generated sheet music.}
        \label{fig:sheetmusic}
\end{figure*}

\autoref{tab:results} shows the evaluation results of our model (Transformer + LM) and the baselines. We selected the recently published semantic dependency network (SDN) as a strong baseline, which already surpassed some methods like LSTM-GAN, SeqGAN, and RelGAN \cite{duan_semantic_2023}. We also implemented the original Transformer as another baseline. The BLEU and ROUGE metrics are slightly worse for the proposed model, however, the difference is insignificant enough to judge that our approach stays relatively close to ground truth in terms of the modeled syllable distribution. In the subjective evaluation in the following sections, and in the generated lyrics from \autoref{sec:gen}, we show that objective metrics can be misleading when evaluating models on a creative task. Examples of generated lyrics accompanied by the input melody are shown in \autoref{fig:sheetmusic}, which show that the lyrics generated by our model can better capture the characteristics of musical lyrics. More generated lyrics by using the proposed methods compared with the baseline model can be seen in \autoref{sec:gen}.


\subsection{ChatGPT evaluation}
\label{sec:chatgpt}

Due to the above-mentioned limitations of objective metrics, we proposed to evaluate the quality and correctness of generated lyrics via Large Language Models (LLMs), since they are objective and have a vast linguistic knowledge. It should be noted that our method only evaluates the texts of lyrics, without considering how well they fit the given melodies. Although feeding symbolic melodies could potentially strain the capabilities of LLMs, it is an approach worth exploring in future work.

We asked the GPT-3 \cite{brown2020language} to evaluate our generated lyrics. After experimenting with the prompts, we proposed the following prompts to let ChatGPT do the evaluation tasks. 

\begin{quote}
    \textit{I will send you three sets of generated candidate lyrics for 20-note melodies. I want you to evaluate them in terms of naturality, correctness, coherence (staying on topic), originality, and poetic value. Try to give numerical scores to all three candidate methods of lyric generation. I will send them in separate messages, please evaluate them after the third message. Is it clear?}
\end{quote}

By clarifying the task by the prompts, we hope to exploit the well-known strong language ability of ChatGPT. The conversation is available online\footnote{https://chat.openai.com/share/46166c1e-5505-4f74-af3d-3627c905b66c}.

In addition to the aforementioned evaluation session, we informed ChatGPT that the lyrics are syllable-split, lowercase, and without punctuation. This additional information made ChatGPT more aware of the characteristics of our input beyond natural language. The conversation of the second version evaluation can be seen at \footnote{https://chat.openai.com/share/bcfdcac3-b63c-44e2-bb29-c93699eae8f2}.

\begin{table}[htb]
\small
\centering
\begin{tabular}{c|cc|cc|cc}
\toprule
\multirow{2}{*}{Metrics} & \multicolumn{2}{c|}{Ground-truth} & \multicolumn{2}{c|}{Transformer} & \multicolumn{2}{c}{Trans.+LM} \\
\cmidrule{2-7}
& 1st & 2nd & 1st & 2nd & 1st & 2nd \\
\midrule
       Naturality  & 6 & 6 & 3 & 5 & \textbf{4} & \textbf{7} \\
       Correctness & 7 & 7 & 4 & 6 & \textbf{5} & \textbf{8} \\
       Coherence & 5 & 5 & 3 & 4 & 3 & \textbf{6} \\
       Originality & 4 & 4 & 2 & 3 & \textbf{3} & \textbf{5} \\
       Poetic Value & 4 & 5 & 2 & 4 & 2 & \textbf{6} \\
\midrule
       Overall & 5.2 & 5.4 & 2.8 & 4.4 & \textbf{3.4} & \textbf{6.4} \\
\bottomrule
\end{tabular}
\caption{Results of the ChatGPT evaluation of generated lyrics on a scale from 1 to 10.}
\label{tab:gpt}
\end{table}

We show the results from both runs in \autoref{tab:gpt}. In both cases, the proposed method is able to outperform the baseline, and in the second evaluation, it also outperforms the ground truth data. During the first evaluation, the ground truth has the highest values in all the categories, while the proposed method is equal to the baseline in two, and outperforms the baseline in 3 of the categories as indicated in bold. During the second evaluation, the proposed method has the highest values in all of the categories. We argue that by clarifying the characteristics of our text input, ChatGPT focuses more on the correctness and quality of the syllable-level split lyrics, hence giving higher scores on our model. This also verified the effectiveness of our proposed methods with language models.

\subsection{Subjective evaluation}

Subjective evaluation is an important metric for evaluating creative text generation systems, especially for evaluating the fitness between the generated lyrics and input melodies.

\subsubsection{Evaluation of generated lyrics}
\label{sec:eval_1}

\begin{figure}[htbp]
    \centering
    \includegraphics[width=\columnwidth]{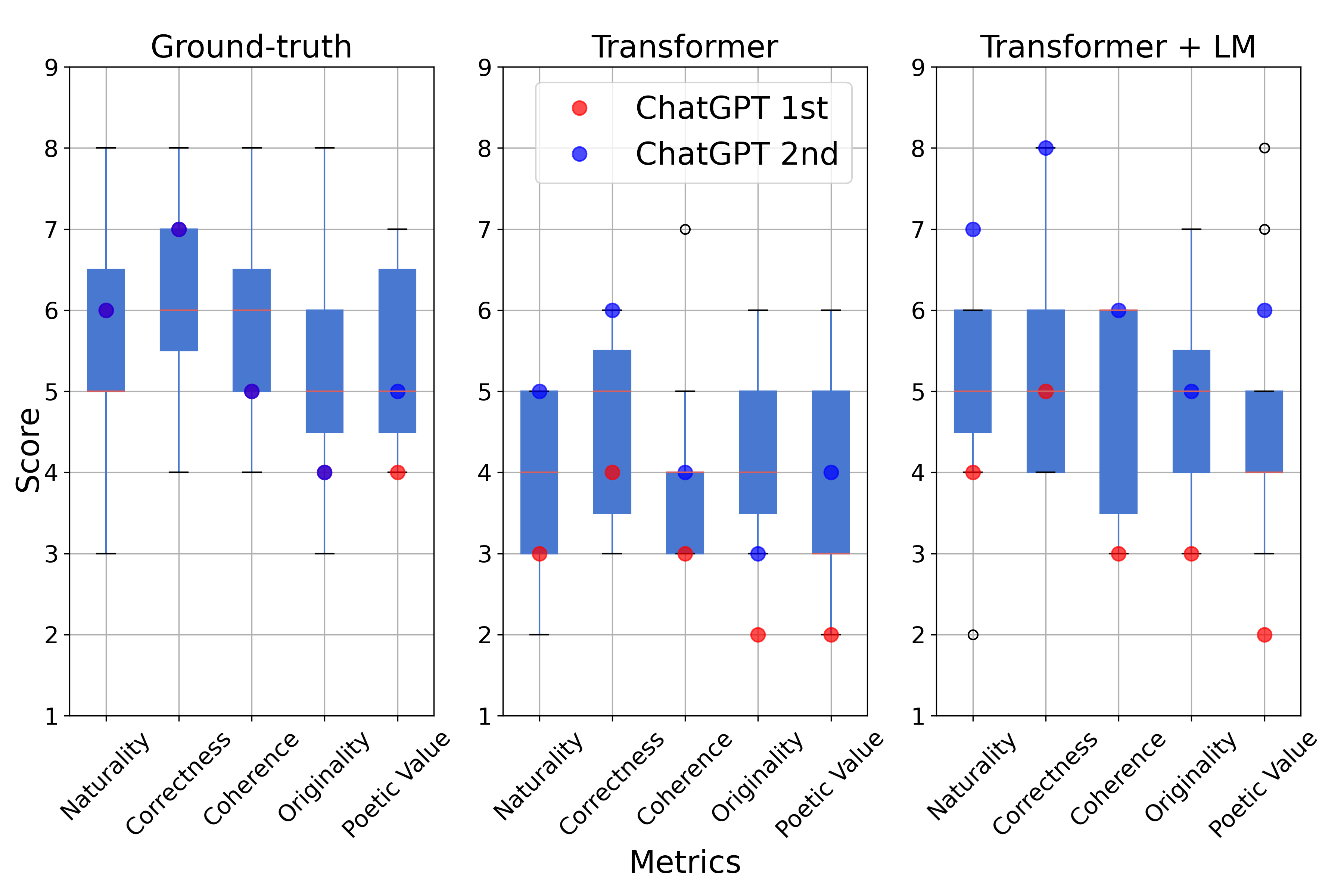}
    \caption{Correlation between ChatGPT-based evaluation and human evaluation of generated lyrics.}
    \label{fig:boxplots}
\end{figure}

We conduct a subjective experiment with the same questions in \autoref{sec:chatgpt} on 11 participants with different levels of musical knowledge to compare human and ChatGPT-based evaluation of texts of generated lyrics. The evaluation results of human participants are visualized via boxplots in \autoref{fig:boxplots}, where we also annotated the ChatGPT-based evaluation results in \autoref{sec:chatgpt} for comparison. We found that human evaluation and ChatGPT-based evaluation show the same general trends among the three methods despite the difference in the numerical scales, where the ground-truth lyrics are rated highest and our model surpasses the Transformer baseline. Moreover, by comparing two sets of ChatGPT-based evaluation results in \autoref{sec:chatgpt}, we found that a more detailed description for ChatGPT about the lyrics to be evaluated is helpful to get the results that are more similar with human evaluation results. However, due to the limited number of participants in our evaluation, it is difficult to perform a thorough correlation analysis. We leave it as future work to conduct a comprehensive analysis with a large number of participants to study the correlation between human and ChatGPT evaluation.

\subsubsection{Evaluation of synthesized music with lyrics and melody}

In addition to the above text-based evaluation of generated lyrics, we performed a subjective evaluation by synthesizing audible samples of our generated lyrics with input melodies and distributing a questionnaire including the audio samples to 11 participants with different levels of musical knowledge. The questionnaire and samples are available at Google Form\footnote{https://forms.gle/RN88Exw3D7H8DjvN7}. We have tried to exclude highly famous songs in the form, to prevent participants from identifying the ground truth hidden reference. The questions used in the subjective evaluation are listed as follows.

\begin{enumerate}
    \item Assess the correctness and coherence of the provided lyrics as natural language, without considering the melody.
    \item What do you think about the creativity and poetic value of the text as song lyrics?
    \item How well do the generated lyrics fit the input melody in terms of rhythm?
    \item How well do the generated lyrics fit the input melody in terms of atmosphere?
\end{enumerate}

\begin{figure}[htbp]
    \centering
    \includegraphics[width=\columnwidth]{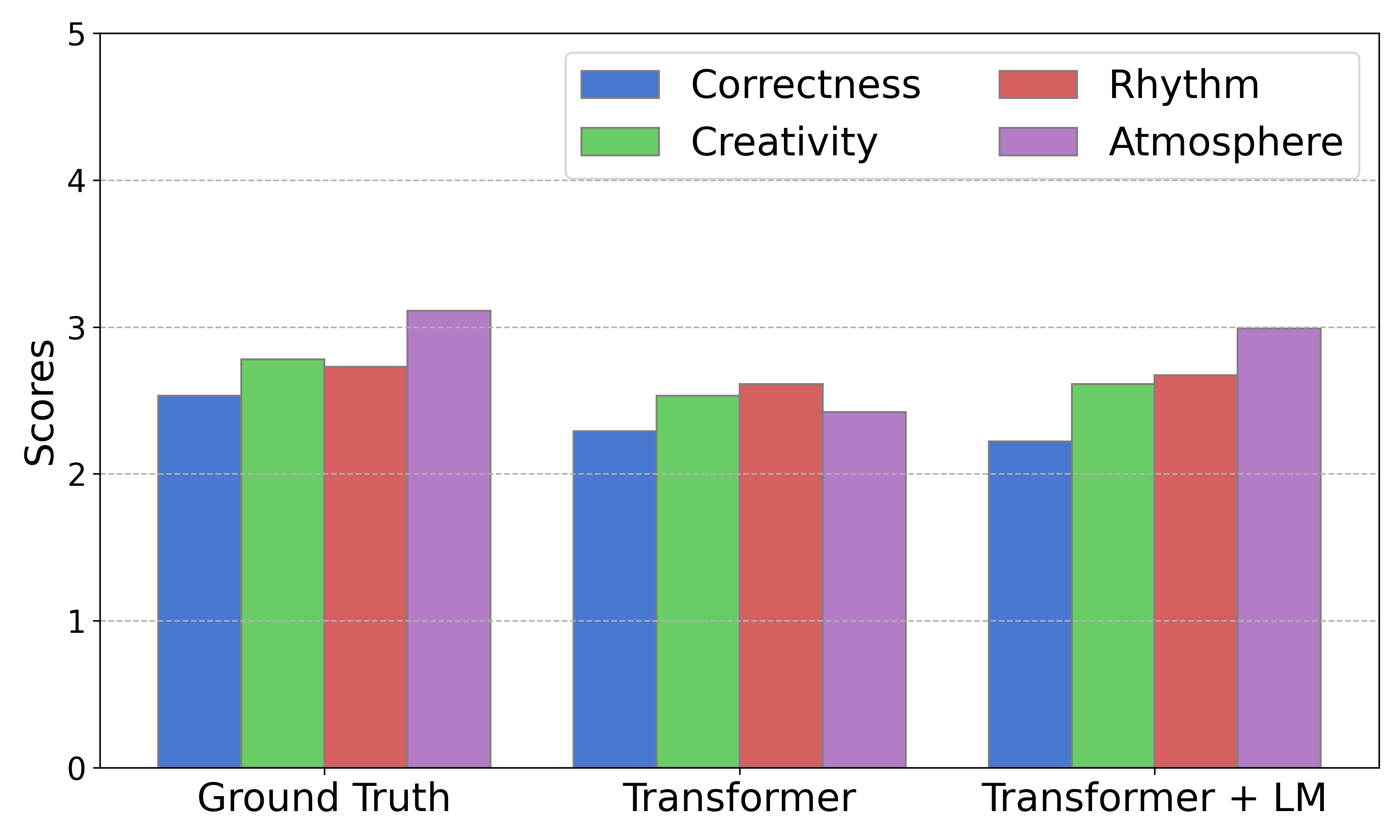}
    \caption{Results of subjective evaluation of lyrics generation from melody.}
    \label{fig:subjective}
\end{figure}

The rating scores are on a 5-point scale (very bad, bad, okay, good, very good). After the subjects finished their questionnaire, we collected the results and calculated the average scores rated for each model. The human evaluation results are shown in \autoref{fig:subjective}. 

Evaluation results show that our proposed model achieves an improvement based on the  Transformer baseline. Also, it is worth mentioning that the potential consistency between human evaluation and ChatGPT evaluation observed in the experiments of \ref{sec:eval_1} makes it promising for future research on ChatGPT-based evaluation, which could be an effective way to improve evaluation efficiency and reduce human resource costs, leveraging the linguistic power of the pre-trained LLMs.

\section{Background and related works}

Lyrics generation has been an active area of research, with various methodologies being proposed over the years. Early efforts in lyrics generation predominantly utilized traditional machine learning methods. For instance, \citet{ramakrishnan_a_automatic_2009} focused on the automatic generation of Tamil lyrics for melodies by predicting the syllable patterns from melodies and subsequently filling the pattern using a corpus.

With the advent of deep learning, there has been a surge in models tailored for automatic lyrics generation. Generating lyrics conditioned on symbolic melody can be thought of as the intersection of creative text generation, and computer music modeling. In both of these areas, recent years have been dominated by deep learning \cite{brown2020language, agostinelli2023musiclm}, leading us to primarily research deep neural networks. \citet{fan_hierarchical_2019} proposed a hierarchical attention-based Seq2Seq model for Chinese lyrics generation that emphasized both word-level and sentence-level contextual information. \citet{lu_syllable-structured_2019} employed RNN encoders for encoding syllable structures and semantic encoding with contextual sentences or input keywords. \citet{wu_hierarchical_2019} introduced a Chinese lyric generation system using an LSTM network to capture the patterns and styles of lyricists. \citet{wang_theme-aware_2019} presented a theme-aware language generation model to enhance the theme-connectivity and coherence of generated paragraphs. Furthermore, \citet{nikolov_rapformer_2020} developed Rapformer, a method that utilizes a Transformer-based denoising autoencoder to reconstruct rap lyrics from extracted content words.

A subset of research has delved deeper into the relationship between lyrics and melodies.  \citet{watanabe_melody-conditioned_2018} proposed a data-driven language model that crafts lyrics for a given input melody. \citet{vechtomova_generation_2020} utilized a bimodal neural network to generate lyrics lines based on short audio clips. \citet{chen_melody-conditioned_2020-2} employed SeqGAN models for syllable-level lyrics generation conditioned on lyrics. \citet{sheng_songmass_2020-3} leveraged unsupervised learning to discern the relationship between lyrics and melodies. \citet{chang_singability-enhanced_2021} introduced a singability-enhanced lyric generator with music style transfer capabilities. \citet{huang_automated_2021} proposed an emotion-based lyrics generation system combining a support vector regression model with a sequence-to-sequence model. \citet{ma_ai-lyricist_2021} presented AI-Lyricist, a system designed to generate vocabulary-constrained lyrics given a MIDI file. \citet{zhang_relyme_2022} and \citet{liu_chipsong_2022} explored methods to enhance the harmony between lyrics and melodies, with the latter focusing on system controllability and interactivity. Lastly, large-scale pre-trained models have also been explored by \cite{rodrigues_lyrics_2022} and Zhang et al. \cite{zhang_youling_2022}.

Many above existing works of lyrics generation are based on word-level sequence generation. In \cite{yu_conditional_2021}, a syllable-level lyrics-melody paired dataset was proposed with an LSTM-GAN model addressing the lyrics-conditioned melody generation problem. Some following works also explored lyrics-to-melody generation problems based on this dataset \cite{yu_lyrics-conditioned_2020,srivastava_melody_2022,duan_interpretable_2022,duan_melody_2023,yu_conditional_2023,zhang_controllable_2023}. However, melody-to-lyrics generation on syllable level is a more difficult task in predicting semantic dependencies among syllable-level, word-level, and sentence-level meaning. A semantic dependency network is proposed in \cite{duan_semantic_2023} to address the degraded text quality in the syllable-level lyrics generation task. In our work, fine-tuning a pre-trained character-level language model is proposed to help the syllable-level melody-to-lyrics Transformer to generate lyrics with better grammar correctness and semantic meaning.

\section{Conclusion}

In this work, we proposed a method to enhance the predictions of a syllable-level melody-conditioned lyrics generation Transformer, which utilizes pre-trained character-level language models fine-tuned on lyrics data. We propose a method for creating a dataset tailored to fine-tune the character-level language model for refining syllable-level semantic meanings. Moreover, we present an algorithm for re-ranking candidate tokens during the beam search procedure.

We prove that our syllable-level refinement leads to improved naturality, correctness, and coherence of lyrics, while maintaining them tightly related to the conditioning melodies via the use of the encoder-decoder architecture. In future work, we plan to work on pre-training a syllable-level language model on a large data corpus, and then fine-tuning it, as well as exploring fine-tuning character-level language models for the task of lyrics-conditioned melody generation.

\section{Limitations}

There are several limitations in the current work and directions for future research:

\begin{enumerate}
    \item Incorporating melody information for ChatGPT evaluation: While our current ChatGPT-based evaluation focuses on the linguistic quality of the generated lyrics, future work could explore ways to provide melody context to ChatGPT, allowing it to evaluate the fit between lyrics and melody.
    \item Expanding the dataset: Our current dataset, though substantial, is limited in its diversity. Gathering more diverse melody-lyrics pairs can further enhance the generalization capabilities of the model.
    \item Exploring other pre-trained models: While we used the CANINE model in our experiments, other character-level or subword-level models could be explored to see if they offer any advantages in this task.
    \item End-to-end training: Instead of a two-step process (Transformer generation followed by language model re-ranking), an end-to-end training approach where both models are jointly trained could be explored.
    \item Risks: It is possible that our method can be utilized to predict lyrics when given melodies. Therefore, it could potentially be leveraged for fake music generation. We will restrict the usage of our method and share our model with the AI community to contribute to the reliability of AI music generation.
\end{enumerate}

\bibliography{Anthology, custom}

\begin{thebibliography}{32}
\expandafter\ifx\csname natexlab\endcsname\relax\def\natexlab#1{#1}\fi

\bibitem[{Agostinelli et~al.(2023)Agostinelli, Denk, Borsos, Engel, Verzetti, Caillon, Huang, Jansen, Roberts, Tagliasacchi, Sharifi, Zeghidour, and Frank}]{agostinelli2023musiclm}
Andrea Agostinelli, Timo~I. Denk, Zalán Borsos, Jesse Engel, Mauro Verzetti, Antoine Caillon, Qingqing Huang, Aren Jansen, Adam Roberts, Marco Tagliasacchi, Matt Sharifi, Neil Zeghidour, and Christian Frank. 2023.
\newblock \href {http://arxiv.org/abs/2301.11325} {Musiclm: Generating music from text}.

\bibitem[{Brown et~al.(2020)Brown, Mann, Ryder, Subbiah, Kaplan, Dhariwal, Neelakantan, Shyam, Sastry, Askell, Agarwal, Herbert-Voss, Krueger, Henighan, Child, Ramesh, Ziegler, Wu, Winter, Hesse, Chen, Sigler, Litwin, Gray, Chess, Clark, Berner, McCandlish, Radford, Sutskever, and Amodei}]{brown2020language}
Tom~B. Brown, Benjamin Mann, Nick Ryder, Melanie Subbiah, Jared Kaplan, Prafulla Dhariwal, Arvind Neelakantan, Pranav Shyam, Girish Sastry, Amanda Askell, Sandhini Agarwal, Ariel Herbert-Voss, Gretchen Krueger, Tom Henighan, Rewon Child, Aditya Ramesh, Daniel~M. Ziegler, Jeffrey Wu, Clemens Winter, Christopher Hesse, Mark Chen, Eric Sigler, Mateusz Litwin, Scott Gray, Benjamin Chess, Jack Clark, Christopher Berner, Sam McCandlish, Alec Radford, Ilya Sutskever, and Dario Amodei. 2020.
\newblock \href {http://arxiv.org/abs/2005.14165} {Language models are few-shot learners}.

\bibitem[{B{\"u}hler et~al.(2005)B{\"u}hler, Minker, and Elciyanti}]{Bhler2005UsingLM}
Dirk B{\"u}hler, Wolfgang Minker, and Artha Elciyanti. 2005.
\newblock Using language modelling to integrate speech recognition with a flat semantic analysis.
\newblock In \emph{SIGDIAL Conferences}.

\bibitem[{Chang et~al.(2021)Chang, Hung, and Lin}]{chang_singability-enhanced_2021}
Jia-Wei Chang, Jason~C. Hung, and Kuan-Cheng Lin. 2021.
\newblock \href {https://doi.org/10.1016/j.comcom.2021.01.002} {Singability-enhanced lyric generator with music style transfer}.
\newblock \emph{Computer Communications}, 168:33--53.

\bibitem[{Chen and Lerch(2020)}]{chen_melody-conditioned_2020-2}
Yihao Chen and Alexander Lerch. 2020.
\newblock \href {https://doi.org/10.1109/ISM.2020.00040} {Melody-{{Conditioned Lyrics Generation}} with {{SeqGANs}}}.
\newblock In \emph{2020 {{IEEE International Symposium}} on {{Multimedia}} ({{ISM}})}, pages 189--196.

\bibitem[{Clark et~al.(2022)Clark, Garrette, Turc, and Wieting}]{Clark_2022}
Jonathan~H. Clark, Dan Garrette, Iulia Turc, and John Wieting. 2022.
\newblock \href {https://doi.org/10.1162/tacl_a_00448} {Canine: Pre-training an efficient tokenization-free encoder for language representation}.
\newblock \emph{Transactions of the Association for Computational Linguistics}, 10:73--91.

\bibitem[{Duan et~al.(2023{\natexlab{a}})Duan, Yu, and Oyama}]{duan_semantic_2023}
Wei Duan, Yi~Yu, and Keizo Oyama. 2023{\natexlab{a}}.
\newblock \href {https://doi.org/10.1007/s00521-023-09282-6} {Semantic dependency network for lyrics generation from melody}.
\newblock \emph{Neural Computing and Applications}.

\bibitem[{Duan et~al.(2023{\natexlab{b}})Duan, Yu, Zhang, Tang, Li, and Oyama}]{duan_melody_2023}
Wei Duan, Yi~Yu, Xulong Zhang, Suhua Tang, Wei Li, and Keizo Oyama. 2023{\natexlab{b}}.
\newblock \href {https://doi.org/10.1145/3572031} {Melody {{Generation}} from {{Lyrics}} with {{Local Interpretability}}}.
\newblock \emph{ACM Transactions on Multimedia Computing, Communications, and Applications}, 19(3):124:1--124:21.

\bibitem[{Duan et~al.(2022)Duan, Zhang, Yu, and Oyama}]{duan_interpretable_2022}
Wei Duan, Zhe Zhang, Yi~Yu, and Keizo Oyama. 2022.
\newblock \href {https://doi.org/10.1145/3503161.3547742} {Interpretable {{Melody Generation}} from {{Lyrics}} with {{Discrete-Valued Adversarial Training}}}.
\newblock In \emph{Proceedings of the 30th {{ACM International Conference}} on {{Multimedia}}}, pages 6973--6975.

\bibitem[{Fan et~al.(2019)Fan, Wang, Zhuang, Wang, and Xiao}]{fan_hierarchical_2019}
Haoshen Fan, Jie Wang, Bojin Zhuang, Shaojun Wang, and Jing Xiao. 2019.
\newblock \href {https://doi.org/10.1007/978-3-030-29894-4_23} {A {{Hierarchical Attention Based Seq2Seq Model}} for {{Chinese Lyrics Generation}}}.
\newblock In \emph{{{PRICAI}} 2019: {{Trends}} in {{Artificial Intelligence}}}, pages 279--288.

\bibitem[{Huang and You(2021)}]{huang_automated_2021}
Yin-Fu Huang and Kai-Cheng You. 2021.
\newblock \href {https://doi.org/10.1109/ACCESS.2021.3095964} {Automated {{Generation}} of {{Chinese Lyrics Based}} on {{Melody Emotions}}}.
\newblock \emph{IEEE Access}, 9:98060--98071.

\bibitem[{Liu et~al.(2022)Liu, Han, Liu, Peng, Zhang, Wang, and Ruan}]{liu_chipsong_2022}
Nayu Liu, Wenjing Han, Guangcan Liu, Da~Peng, Ran Zhang, Xiaorui Wang, and Huabin Ruan. 2022.
\newblock \href {https://doi.org/10.18653/v1/2022.in2writing-1.13} {{{ChipSong}}: {{A Controllable Lyric Generation System}} for {{Chinese Popular Song}}}.
\newblock In \emph{Proceedings of the {{First Workshop}} on {{Intelligent}} and {{Interactive Writing Assistants}} ({{In2Writing}} 2022)}, pages 85--95.

\bibitem[{Lu et~al.(2019)Lu, Wang, Zhuang, Wang, and Xiao}]{lu_syllable-structured_2019}
Xu~Lu, Jie Wang, Bojin Zhuang, Shaojun Wang, and Jing Xiao. 2019.
\newblock \href {https://doi.org/10.1007/978-3-030-29894-4_20} {A {{Syllable-Structured}}, {{Contextually-Based Conditionally Generation}} of {{Chinese Lyrics}}}.
\newblock In \emph{{{PRICAI}} 2019: {{Trends}} in {{Artificial Intelligence}}}, pages 257--265.

\bibitem[{Ma et~al.(2021)Ma, Wang, Kan, and Lee}]{ma_ai-lyricist_2021}
Xichu Ma, Ye~Wang, Min-Yen Kan, and Wee~Sun Lee. 2021.
\newblock \href {https://doi.org/10.1145/3474085.3475502} {{{AI-Lyricist}}: {{Generating Music}} and {{Vocabulary Constrained Lyrics}}}.
\newblock In \emph{Proceedings of the 29th {{ACM International Conference}} on {{Multimedia}}}, pages 1002--1011.

\bibitem[{Nikolov et~al.(2020)Nikolov, Malmi, Northcutt, and Parisi}]{nikolov_rapformer_2020}
Nikola~I. Nikolov, Eric Malmi, Curtis~G. Northcutt, and Loreto Parisi. 2020.
\newblock \href {https://doi.org/10.48550/arXiv.2004.03965} {Rapformer: {{Conditional Rap Lyrics Generation}} with {{Denoising Autoencoders}}}.

\bibitem[{Ramakrishnan~A et~al.(2009)Ramakrishnan~A, Kuppan, and Lalitha~Devi}]{ramakrishnan_a_automatic_2009}
Ananth Ramakrishnan~A, Sankar Kuppan, and Sobha Lalitha~Devi. 2009.
\newblock Automatic {{Generation}} of {{Tamil Lyrics}} for {{Melodies}}.
\newblock In \emph{Proceedings of the {{Workshop}} on {{Computational Approaches}} to {{Linguistic Creativity}}}, pages 40--46.

\bibitem[{Rodrigues et~al.(2022)Rodrigues, Oliveira, Moreira, and Possi}]{rodrigues_lyrics_2022}
Matheus~Augusto Rodrigues, Alcione Oliveira, Alexandra Moreira, and Maurilio Possi. 2022.
\newblock \href {https://doi.org/10.32473/flairs.v35i.130607} {Lyrics {{Generation}} supported by {{Pre-trained Models}}}.
\newblock \emph{The International FLAIRS Conference Proceedings}, 35.

\bibitem[{Sheng et~al.(2020)Sheng, Song, Tan, Ren, Ye, Zhang, and Qin}]{sheng_songmass_2020-3}
Zhonghao Sheng, Kaitao Song, Xu~Tan, Yi~Ren, Wei Ye, Shikun Zhang, and Tao Qin. 2020.
\newblock \href {https://doi.org/10.48550/arXiv.2012.05168} {{{SongMASS}}: {{Automatic Song Writing}} with {{Pre-training}} and {{Alignment Constraint}}}.

\bibitem[{Srivastava et~al.(2022)Srivastava, Duan, Shah, Wu, Tang, Li, and Yu}]{srivastava_melody_2022}
Abhishek Srivastava, Wei Duan, Rajiv~Ratn Shah, Jianming Wu, Suhua Tang, Wei Li, and Yi~Yu. 2022.
\newblock \href {https://doi.org/10.1007/978-3-030-98358-1_45} {Melody {{Generation}} from {{Lyrics Using Three Branch Conditional LSTM-GAN}}}.
\newblock In \emph{{{MultiMedia Modeling}}}, pages 569--581.

\bibitem[{Vaswani et~al.(2017)Vaswani, Shazeer, Parmar, Uszkoreit, Jones, Gomez, Kaiser, and Polosukhin}]{vaswani2017attention}
Ashish Vaswani, Noam Shazeer, Niki Parmar, Jakob Uszkoreit, Llion Jones, Aidan~N. Gomez, Lukasz Kaiser, and Illia Polosukhin. 2017.
\newblock \href {http://arxiv.org/abs/1706.03762} {Attention is all you need}.

\bibitem[{Vechtomova et~al.(2020)Vechtomova, Sahu, and Kumar}]{vechtomova_generation_2020}
Olga Vechtomova, Gaurav Sahu, and Dhruv Kumar. 2020.
\newblock \href {https://doi.org/10.48550/arXiv.2009.14375} {Generation of lyrics lines conditioned on music audio clips}.

\bibitem[{Wang and Zhao(2019)}]{wang_theme-aware_2019}
Jie Wang and Xinyan Zhao. 2019.
\newblock \href {https://doi.org/10.48550/arXiv.1906.02134} {Theme-aware generation model for chinese lyrics}.

\bibitem[{Wang et~al.(2021)Wang, Meng, Sun, Wu, Ouyang, Yan, Zhang, and Li}]{wang2021modeling}
Shuhe Wang, Yuxian Meng, Xiaofei Sun, Fei Wu, Rongbin Ouyang, Rui Yan, Tianwei Zhang, and Jiwei Li. 2021.
\newblock \href {http://arxiv.org/abs/2105.14445} {Modeling text-visual mutual dependency for multi-modal dialog generation}.

\bibitem[{Watanabe et~al.(2018)Watanabe, Matsubayashi, Fukayama, Goto, Inui, and Nakano}]{watanabe_melody-conditioned_2018}
Kento Watanabe, Yuichiroh Matsubayashi, Satoru Fukayama, Masataka Goto, Kentaro Inui, and Tomoyasu Nakano. 2018.
\newblock \href {https://doi.org/10.18653/v1/N18-1015} {A {{Melody-Conditioned Lyrics Language Model}}}.
\newblock In \emph{Proceedings of the 2018 {{Conference}} of the {{North American Chapter}} of the {{Association}} for {{Computational Linguistics}}: {{Human Language Technologies}}, {{Volume}} 1 ({{Long Papers}})}, pages 163--172.

\bibitem[{Wolf et~al.(2019)Wolf, Debut, Sanh, Chaumond, Delangue, Moi, Cistac, Rault, Louf, Funtowicz, Davison, Shleifer, von Platen, Ma, Jernite, Plu, Xu, Scao, Gugger, Drame, Lhoest, and Rush}]{huggingface-https://doi.org/10.48550/arxiv.1910.03771}
Thomas Wolf, Lysandre Debut, Victor Sanh, Julien Chaumond, Clement Delangue, Anthony Moi, Pierric Cistac, Tim Rault, Rémi Louf, Morgan Funtowicz, Joe Davison, Sam Shleifer, Patrick von Platen, Clara Ma, Yacine Jernite, Julien Plu, Canwen Xu, Teven~Le Scao, Sylvain Gugger, Mariama Drame, Quentin Lhoest, and Alexander~M. Rush. 2019.
\newblock \href {https://doi.org/10.48550/ARXIV.1910.03771} {Huggingface's transformers: State-of-the-art natural language processing}.

\bibitem[{Wu et~al.(2019)Wu, Du, Guo, and Fujita}]{wu_hierarchical_2019}
Xing Wu, Zhikang Du, Yike Guo, and Hamido Fujita. 2019.
\newblock \href {https://doi.org/10.1007/s10489-018-1206-2} {Hierarchical attention based long short-term memory for {{Chinese}} lyric generation}.
\newblock \emph{Applied Intelligence}, 49(1):44--52.

\bibitem[{Yu et~al.(2020)Yu, Harsco{\"e}t, Canales, Reddy~M, Tang, and Jiang}]{yu_lyrics-conditioned_2020}
Yi~Yu, Florian Harsco{\"e}t, Simon Canales, Gurunath Reddy~M, Suhua Tang, and Junjun Jiang. 2020.
\newblock \href {https://doi.org/10.1007/978-3-030-37734-2_58} {Lyrics-{{Conditioned Neural Melody Generation}}}.
\newblock In \emph{{{MultiMedia Modeling}}: 26th {{International Conference}}, {{MMM}} 2020, {{Daejeon}}, {{South Korea}}, {{January}} 5\textendash 8, 2020, {{Proceedings}}, {{Part II}}}, pages 709--714.

\bibitem[{Yu et~al.(2021)Yu, Srivastava, and Canales}]{yu_conditional_2021}
Yi~Yu, Abhishek Srivastava, and Simon Canales. 2021.
\newblock \href {https://doi.org/10.1145/3424116} {Conditional {{LSTM-GAN}} for {{Melody Generation}} from {{Lyrics}}}.
\newblock \emph{ACM Transactions on Multimedia Computing, Communications, and Applications}, 17(1):35:1--35:20.

\bibitem[{Yu et~al.(2023)Yu, Zhang, Duan, Srivastava, Shah, and Ren}]{yu_conditional_2023}
Yi~Yu, Zhe Zhang, Wei Duan, Abhishek Srivastava, Rajiv Shah, and Yi~Ren. 2023.
\newblock \href {https://doi.org/10.1007/s00521-022-07863-5} {Conditional hybrid {{GAN}} for melody generation from lyrics}.
\newblock \emph{Neural Computing and Applications}, 35(4):3191--3202.

\bibitem[{Zhang et~al.(2022{\natexlab{a}})Zhang, Chang, Wu, Tan, Qin, Liu, and Zhang}]{zhang_relyme_2022}
Chen Zhang, Luchin Chang, Songruoyao Wu, Xu~Tan, Tao Qin, Tie-Yan Liu, and Kejun Zhang. 2022{\natexlab{a}}.
\newblock \href {https://doi.org/10.1145/3503161.3548357} {{{ReLyMe}}: {{Improving Lyric-to-Melody Generation}} by {{Incorporating Lyric-Melody Relationships}}}.
\newblock In \emph{Proceedings of the 30th {{ACM International Conference}} on {{Multimedia}}}, pages 1047--1056.

\bibitem[{Zhang et~al.(2022{\natexlab{b}})Zhang, Mao, Li, Jiang, Chen, Hu, Xi, Fan, and Huang}]{zhang_youling_2022}
Rongsheng Zhang, Xiaoxi Mao, Le~Li, Lin Jiang, Lin Chen, Zhiwei Hu, Yadong Xi, Changjie Fan, and Minlie Huang. 2022{\natexlab{b}}.
\newblock \href {https://doi.org/10.48550/arXiv.2201.06724} {Youling: An {{AI-Assisted Lyrics Creation System}}}.

\bibitem[{Zhang et~al.(2023)Zhang, Yu, and Takasu}]{zhang_controllable_2023}
Zhe Zhang, Yi~Yu, and Atsuhiro Takasu. 2023.
\newblock \href {https://doi.org/10.1007/s00521-023-08728-1} {Controllable lyrics-to-melody generation}.
\newblock \emph{Neural Computing and Applications}, 35(27):19805--19819.

\end{thebibliography}

\appendix

\section{Generated lyrics}
\label{sec:gen}

\begin{table*}[htbp]
\centering
\small
\begin{tabularx}{\textwidth}{X|X|X}
\toprule
Ground truth & Transformer & Transformer + LM \\ 
\midrule
how minus cule is any light if it light you breaking up the fold for your love & when it takes more than i met you the sub way the pow er of the stars & not the way you bet ter than you ev er seen it when you need some thing  \\ \\
i need to know the way to feel to keep me sat is fied & i know i be lieve i can give you the way it got to & i know i believe in love with you to mor row bless me soon \\ \\
in their mas que rade no the out to get you & you got ta o in what the you used to life & you got ta treat me to may be un der stand you \\ \\
and i touched her on the sle e ve she rec og nize the face at first & and i be lieve i can fol low you know i must have known it ea sy & but i be lieve i can fol low you know i have to face it of us \\ \\
da la da da la da da drift a way fade a way lit tle tin god dess & da da da la da da da da da da la da da da da da da da & we can give this world to ge ther and we are not so da da da da da  \\ \\
from mem phis ten nes see her home is on the south side high up on a ridge & for get a no ther way you real ly need to know now when it feels like you & for get a no ther way you real ly need to know now when it feels so hard \\ \\
went crash boom bang the whole rhy thm sec tion was the pur ple gang rock & must have been ran ing to the an swer to we got no thing no thing & must have been talk ing to the an swer i wan na live for some thing \\ \\
you take mur der on the & in the wings of the ri ver & in the wings of the ri ver \\ \\
with you and the lit tle days and party joints do now just miss ing you how i wish & a gain why i come a gain why i must be my su per to me smil ing like & a gain why i must re mem ber the sun shine fills my head with me and she stings  \\ \\
i want to break free i want i want i want i want to break free to break free & i ne ver on ly know i on ly know who i am i was born on a wall & i should be here i am i ne ver seen your horse and i know what i feel inside  \\ \\
and it it makes me me sad for the ly walked that road for so now i know that the & i stand the ground and i stand the fire my friend and i need a rai ny roads i need & i stand the ground and i stand the fire my friend some times i need a rai ny roads run  \\ \\
do ing do wop do we were in the with our blue suede shoes & an y li ons they say that you were a life of your life & they know what they think that they were six teen your world a bout  \\ \\
i got my first real bought it at the played it till my fin gers bled & and she looks so hard to un der stand that she comes the game and they & and she looks so hard to un der stand the word and they come to town  \\ \\
to it mad bur ning mad it it mad ni ght the beat to the beat to the beat & to you know gon na be a and i your to be doing the the the be oh the & to night gon na be out of the night babe cos i will be a called love grow when \\ \\
get down and move it a round hey love need girl you tell if feel too in hour & what i hea ven no bod y no bod y wants what i heard you a ny thing & your bod y call me your bod y sis ter su per star hol low and too much \\ \\
she rush es out to hold him thank ful a live but on the wind and rain & and by the way you come a lit tle bit more you get a lit tle clos & you can say a bout my love for you to day and you get a feel ing \\ \\
must be how could so much love be in side of you whoa oh & on the run ning on the run ning to be with you to town & on the road got to be shin ing on the streets of the town  \\ \\
high out side your door late at night when not sleep ing and moon light falls a cross your floor & why do i have to die why we won der where it was the rain bow is fall ing down & why do i have to die why we won der where it was the rain bow is fall ing down \\ \\
ma ha mm ma ha ha ha ha ha ha the world & she said got me love for me but each oth er day & she said got me love for me but each oth er day  \\ \\
love has tak en life time child girl you know you are the nic est thing love your rap & sex bomb and can you feel the one smile you know you smile you smile i want to cry & sex bomb and smile you take the mo ney sex bomb and smile you know talk to get back \\ \\
the glo ries of his righ teous ness and won ders of his love and won ders of his love & un der stand why mark and if on ly i say this is ach ing you if i do this & un til this day i wear my heart and try to bring me out of mind if i should let  \\ 
\bottomrule
\end{tabularx} 
    \caption{Comparison of generated lyrics.}
    \label{tab:lyr1}
\end{table*}

\end{document}